\documentclass{article}

\usepackage[final]{neurips_2019}

\usepackage[utf8]{inputenc} 
\usepackage[T1]{fontenc}    
\usepackage{hyperref}       
\usepackage{url}            
\usepackage{booktabs}       
\usepackage{amsfonts}       
\usepackage{amsmath}
\usepackage{nicefrac}       
\usepackage{microtype}      
\usepackage{tikz}
\usepackage{graphicx}
\usepackage{amsthm}
\usepackage{subfig}
\usepackage{float}
 
\theoremstyle{definition}
\newtheorem*{definition}{Definition}

\usetikzlibrary{graphs,arrows,decorations.pathmorphing,backgrounds,positioning,fit,petri,arrows,decorations.markings}

\newcommand\margin {12}     
\newcommand\n {8}           
\newcommand\radius {2cm}    
\newcommand\setspace{1cm}   

\newcommand\nodedistance{12}        
\newcommand\nodesize{12}            
\newcommand\arrowheadsize{2}        
\newcommand\nodefontsize{\large}    

\newcommand\arrowheadsizediag{1.5}

\newcommand{\expect}{\mathbb{E}}
\newcommand{\KL}[2]{\mathcal{D}_{\text{KL}}[#1\,||\,#2]}

\DeclareMathOperator{\Ob}{\text{Ob}}
\DeclareMathOperator{\Hom}{\text{Hom}}

\setcitestyle{numbers,square} 

\title{Category-Learning\\ with Context-Augmented Autoencoder}
\author{Denis Kuzminykh${}^1$, Laida Kushnareva${}^2$, Timofey Grigoryev${}^1$,  Alexander Zatolokin${}^1$ \vspace{2pt}\\
    ${}^1$\textit{Moscow Institute of Physics and Technology}\\
    ${}^2$\textit{Lomonosov Moscow State University}}

\date{}
\begin{document}

\maketitle

\begin{abstract}
Finding an interpretable non-redundant representation of real-world data is one of the key problems in Machine Learning. Biological neural networks are known to solve this problem quite well in unsupervised manner, yet unsupervised artificial neural networks either struggle to do it or require fine tuning for each task individually. We associate this with the fact that a biological brain learns in the context of the relationships between observations, while an artificial network does not. We also notice that, though a naive data augmentation technique can be very useful for supervised learning problems, autoencoders typically fail to generalize transformations from data augmentations. Thus, we believe that providing additional knowledge about relationships between data samples will improve model's capability of finding useful inner data representation. More formally, we consider a dataset not as a manifold, but as a category, where the examples are objects. Two these objects are connected by a morphism, if they actually represent different transformations of the same entity. Following this formalism, we propose a novel method of using data augmentations when training autoencoders. We train a Variational Autoencoder in such a way, that it makes transformation outcome predictable by auxiliary network in terms of the hidden representation. We believe that the classification accuracy of a linear classifier on the learned representation is a good metric to measure its interpretability. In our experiments, present approach outperforms $\beta$-VAE and is comparable with Gaussian-mixture VAE.
\end{abstract}

\smallskip

\section{Introduction}

The performance of machine learning algorithms crucially depends on the representation of data they work with. Relevant for the particular task data representation can significantly improve the performance of machine learning models on this task. As it was shown by Schmidhuber el al.~in \cite{schmidhuber}, Bengio et al.~in \cite{bengio} and I. Higgins et al.~in \cite{higgins}, disentangled (or statistically independent) representations are particularly valuable, because they are useful in a wide range of applied tasks, particularly in the computer vision field.

Advanced supervised machine learning models are known to be able to learn high-abstract relevant non-redundant representations, but they require specific preprocessing and a large amount of labeled data to achieve this. However, labeled data are usually limited and expensive to obtain. That is why the problem of learning good data representation from unlabeled data is important.

The requirements to amount of data required for learning good representation can be softened by using data augmentation technique. Initial dataset can be expanded by applying set of transformations to each training sample, where each of these transformation is known a priori to not affect desired output. However, when using data augmentation, we still need decent initial amount of labeled data, and only labeled data can be used. 

A frequent situation in practice is when large amount of unlabeled samples is available, but labeling them is expensive. While data augmentation helps to significantly improve results in supervised tasks and learn good data representation during training, it usually does not help to learn more meaningful representation when applied naively in unsupervised tasks.

To see the problem, consider an example of CNN autoencoder for digit images, and assume we want it to extract an interpretable compact image representation, such that we can recognize digit from it easily (by linear classifier) regardless of rotation, scale, etc. If we provide our model an image of \textbf{3} and the same image, rotated $45^\circ$, it is really hard to infer relation between these images from training signal (pixel-wise reconstruction loss). In fact, in terms of pixel $L_2$ distance (typically used as a reconstruction loss) \textbf{3} is closer to \textbf{9} than to $45^\circ$ rotated \textbf{3} (contrary to classification task, where the rotation invariance is implicitly encoded in the training signal). Because of this, applying rotation augmentations will help the autoencoder to memorize all rotated variations of digits, but it still fails to generalize image classes over rotations. This case was experimentally investigated in \cite{kuzminykh}. This kind of generalization we want to obtain can be equivalently formulated as disentanglement of class-related part of image description from rotation related part in terms of linear subspaces, i.e. factoring by rotation group.

More generally, when we work with natural images, we want to separate image description space into two subspaces of features: the feature subspace, representing inherent properties of the depicted objects themselves and the feature subspace, representing different kinds of transformations (zooms, rotations, shifts etc). 

One particular approach to such a separation was proposed by D. Kuzminykh et al.~in \cite{kuzminykh}.
By using properties of the Group Equivariant Convolutional Networks \cite{kuzminykh}, the autoencoder with specialized architecture can factor it's internal representation space by group of transformations used by group equivariant convolutions in encoder network, allowing to distill all invariant information. It is shown that such representation can successfully disentangle image classes, making them linearly separable once irrelevant transformation groups are factored out. However, that approach has number of disadvantages, arising from reliance on group equivariant convolutions: group equivariant convolution operation needs to be implemented manually for desired group, which is not always possible or require additional tricks, set of transformations need to be proper algebraic group with known structure, computation and memory costs grow with the group size.

In this paper, we introduce new, more flexible approach, inspired by observations on how biological brains do unsupervised learning. Instead of using specialized architecture to factor particular group of transformations, we change learning process to train encoder to produce more predictable image descriptions. To do so, we go away from training autoencoder on i.i.d. samples, and instead train it on conditionally connected samples, represented as tuples (image, transformation, result).

\subsection{Biological Inspiration}

Biological neural networks are known to learn good data representation in an unsupervised manner. We suppose that one of the main reasons for it is that brain always learns in context. It remembers what it has seen before and tries to predict what will happen next. Particularly, a brain of an animal always knows where the animal is moving at every moment due to the signals it exchanges with muscles.
Incredible importance of such a context was shown in paper \cite{held}. Authors described a biological experiment, which was performed on pairs of newborn kittens at early stage of their post-natal development, which is considered as crucial for their learning. Each pair was placed into an apparatus, shown on figure 1. One of the kittens (active subject, $A$) had relative freedom to move actively around axes $a$, $b$ and $c$. Another one (passive subject, $P$) was fixed inside of gondola, which repeated movements of $A$ automatically, due to mechanical transmission system. Passive subject couldn't influence its own movements. In the same time, both subjects could equally observe their environment.

\begin{figure}[htp]
    \centering
    \includegraphics[width=10cm]{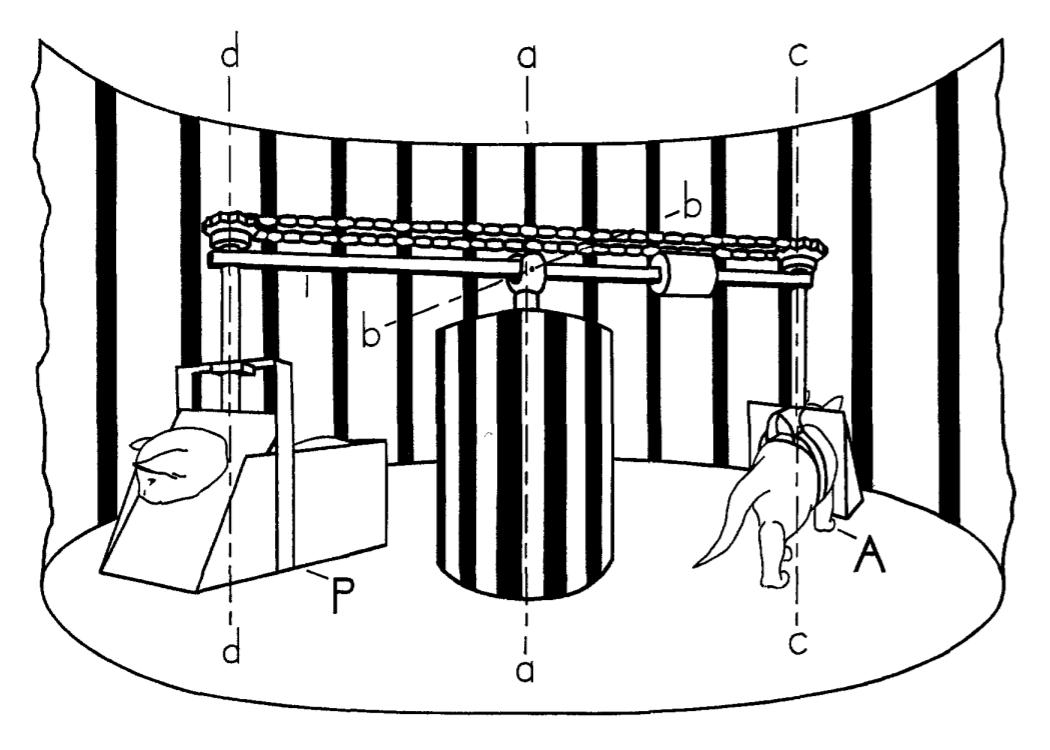}
    \caption{Apparatus for equating motion and consequent visual feedback for an actively moving (A)
and a passive passively moved (P) S. Reprinted from \cite{held}, p.873}
    \label{fig:cats}
\end{figure}

After spending suitable amount of time inside this apparatus, kittens were released, and their responses on visual stimuli were tested. Passive kittens didn't manage to react properly on approaching objects and obstacles, as if they didn't see them properly, despite the fact, that their eyes were healthy. In the same time, active kittens didn't have such a problem. They develop normal response on visual environment. Active kittens were able to learn relationship between their own movements and the visual data they received, and this context turned out to be crucial for them to learn, how to interpret visual data in a meaningful way.

But unsupervised machine learning models usually don't have such a context. Their sources of information are limited to distinct data samples from a training set via the loss function. It doesn't have context that a biological brain has. Autoencoder models ``know'' that there is some compressed data representation and that they ``should'' find it, but found compressed representation isn't necessarily meaningful. Moreover, the task of compressing data can be performed fully without understanding the data structure. For example, JPEG algorithm is able to compress images, but is fully unable to extract useful features from it.

Having in mind the biological analogy, we think, that providing an additional context to autoencoder would let it learn more abstract and meaningful features. Particularly, we want to teach it the link between transformation (movement) and a new picture, which it gets after the movement.

\section{Related Work}

There are several different approaches to the described problem, known so far. Two especially influential of them are $\beta$-VAE \cite{held} and learning factorial codes by predictability minimization \cite{higgins}. Among more recent approaches, we want to highlight Multi-Modal Deep Clustering \cite{mmdc}, Maximizing Mutual Information Across Views \cite{mmiav} and Contrastive Learning \cite{chen}. Let us briefly highlight the insights of these works.

\subsection{Factorial Codes}

Factorial Codes are a particular variation of the lossless encoding. A factorial code of a given data piece is a vector of statistically independent features, which can be decoded into that original piece of data again. In other words, that vector should get rid from statistical dependencies inside the data piece, but save all the information, which it contained, in the same time.

A particular way for unsupervised learning of such a non-redundant representation was introduced in \cite{higgins}. There authors consider an abstract learner, which learns a representation of the data, using some hidden units. Then they assign an adaptive predictor to each hidden (representation) unit of this learner. Each predictor learn to predict the output of the corresponding unit, using the outputs of all other units of the same representational layer. In turn, each representational unit try to minimize its own predictability. This competition encourage each representational unit to learn concepts, statistically independent of those upon which the other units focused.

Authors also demonstrate a particular implementation of this idea, performing experiments with a particular neural network architecture and a learning algorithm for it.

\subsection{\texorpdfstring{$\beta$}--VAE}

Authors of $\beta$-VAE adapted well-known variational autoencoder \cite{VAE} model for learning factorized interpretable data representations. To achieve this, they specify disentangled metric and use it to enhance VAE's loss function in a way so to make it consider an ``entanglement'' as a part of the loss.

This enhanced loss function includes a hyperparameter $\beta$, which balances latent channel capacity and independence constraints with reconstruction accuracy (and which gives the model it's name). The edge case of $\beta = 1$ gives the classical VAE model. The further $\beta$ parameter is from 1, the more the ``entanglement'' part of the loss matters.

\subsection{Multi-Modal Deep Clustering}

In \cite{mmdc} a new algorithm for training a CNN-based neural network for natural images clustering is proposed. This algorithm combines main, unsupervised task of "noise as target" encoding with auxiliary task of rotation prediction, in order to obtain a representation with meaningful, clearly defined clusters.

"Noise as target" method aligns embeddings of images with target points sampled from a Gaussian Mixture. For this purpose it utilizes loss function based on the distance between image embedding and a closest target point. The additional task of rotation prediction encourages rotations of the same image to map into close vectors in the representation layer.

\subsection{Maximizing Mutual Information Across Views}

Authors of \cite{mmiav} encourage their model to maximize mutual information between features extracted from independently-augmented copies of the same image. The purpose of using such augmentations in this particular work is to create multiple views of a shared context, and then to force a neural network to extract the most informative features, shared by all these views.

\subsection{Simple Framework for Contrastive Learning}

Authors of SimCLR \cite{chen} also aim to extract transformation-independent representations of visual data. They achieve this goal by combining original architecture, composed of encoder and an auxiliary network, with a suitable dataset augmentation. It's easy to see that our approaches share much in common; however, they have deep differences as well. We will point out these differences below, right after giving brief outline of the Contrastive Learning framework itself.

In SimCLR, convolutional enconder takes two different transformations of the same image as independent inputs and learns representation of each one inside its latent space. In turn auxiliary fully-connected network called the ``projection head'' learns to map (project) representations corresponding to transformations of the same images into vectors, close in terms of the cosine similarity. Both networks train together by minimizing normalized temperature-scaled cross entropy loss through gradient descent (see the original paper \cite{chen} for more details).

In our work auxiliary network plays another role: it learns transformations instead of projections. Besides, \cite{chen} operates with representations, which drop information about transformations, while our model aims to save information about transformations in a disentangled manner. It is achieved by basing our model on an autoencoder instead of just encoder.

\bigskip

All these results have shown the way in which the resulting data representation is useful in a variety of tasks, especially in the computer vision field. However, first two papers attempted to approach the problem as solely a part of a compression task. Recognizing the huge importance of their results, we nevertheless admit that those techniques are not sufficient to learn good representation of more complicated datasets. In turn, the last three papers revolve around extraction of transformation invariants, but without keeping information about transformations themselves in the latent layer. So, despite the fact that such representations are proved to be useful, they are not decodable.

Our model, on the other hand, uses these invariants for disentangling learned representation, keeping it decodable. To achieve this, we augment the underlying VAE model in such a way, that it not only solves the compression task, but also tries to predict certain objects transformations. We see it as a part of an important learning context discussed above, which $\beta$-VAE and factorial codes models lack by themselves. We also provide formal mathematical description of our idea, using the Category Theory language as well as the Bayesian inference, justifying our architectural solutions.

\section{Model Description}

\subsection{Categorical Background}

\begin{figure}
    \centering    
    \nodefontsize
      
    \begin{tikzpicture}
        \foreach \s in {1,...,\n}
        {
           \node[draw, rectangle, rotate=(\s*360/\n)] at ({1.5*360/\n * (\s - 1)}:\radius - \s*1.5 mm + 5mm) {$3$};
        }
        \node at (0, 0) (Set) {$Set$};
    \end{tikzpicture}
    \hspace{\setspace}
    \begin{tikzpicture}
        \foreach \s in {1,...,\n}
        {
           \node[draw, rectangle, rotate=(\s*360/\n)] at ({360/\n * (\s - 1)}:\radius) {$3$};
           \draw[->, >=latex] ({360/\n * (\s - 1)+\margin}:\radius) 
             arc ({360/\n * (\s - 1)+\margin}:{360/\n * (\s)-\margin}:\radius);
        }
        \node at (0, 0) (Cat) {$Category$};
    \end{tikzpicture}
    
    \caption{Set~vs.~Category}
    \label{set_vs_category}
    \normalsize
\end{figure}
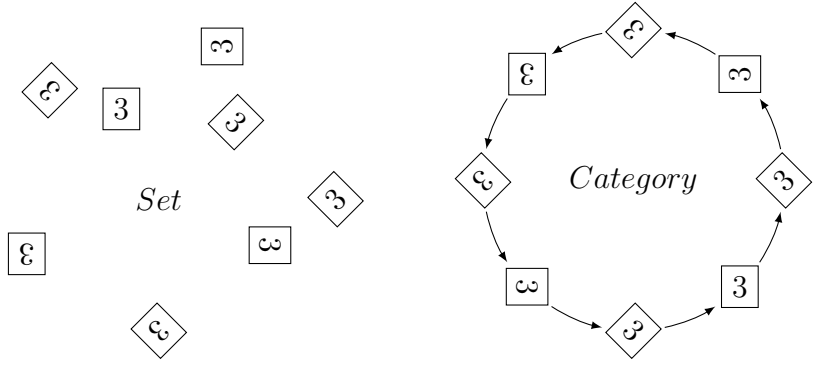

Since we want to take into account the relationships between samples in our dataset, we suggest describing our dataset in the terms of the Category Theory. To begin, we recall the base definition of a category.

\smallskip

\begin{definition}
    A category $\mathcal{C}$ is an aggregate of two classes --- the class of objects $\Ob(\mathcal{C})$ and the class of morphisms $\Hom(\mathcal{C})$ with the following properties:
    \begin{enumerate}
    \item Each pair of objects $a, b \in \Ob(\mathcal{C})$ corresponds to a set $H_\mathcal{C}(a, b) \subset \Hom(\mathcal{C})$ of morphisms.
    
    \item Each morphism $\alpha \in \Hom(\mathcal{C})$ belongs to one and only one set $H_\mathcal{C}(x, y)$, where $x$ and $y$ are some objects from $\Ob(\mathcal{C})$.
    
    \item There is a composition operation defined in the class $\Hom(\mathcal{C})$. The composition of $\alpha \in H_\mathcal{C}(a, b)$ and $\beta \in H_\mathcal{C}(b, c)$ gives a morphism $\beta\alpha \in H_\mathcal{C}(a, c)$. This operation is associative, i.e. $\gamma(\beta\alpha) = (\gamma\beta)\alpha$ for any three morphisms $\alpha \in H_\mathcal{C}(a, b),\, \beta \in H_\mathcal{C}(b, c),\, \gamma \in H_\mathcal{C}(c, d)$.
    
    \item Each set $H_\mathcal{C}(a, a)$ contains a special morphism $1_a$, such that for any morphisms $\alpha \in H_\mathcal{C}(x, a)$ and $\beta \in H_\mathcal{C}(a, y)$ $1_a \alpha = \alpha$ and $\beta 1_a = \beta$. This special morphism is called the identity morphism of object $a$.
    \end{enumerate}
\end{definition}

\smallskip

For more detailed explanation, see \cite{category}.

\smallskip

Returning to our situation, let's consider the set $\mathbf{X}$ of observations and the set $\mathbf{A}$ of allowed transformations between these observations. For example, $\mathbf{X}$ can be the set of some natural images, and $\mathbf{A}$ --- the set of possible rotations, shifts and rescalings of each object. Now, we introduce a category $\mathcal{X}$ in a following way:

\begin{itemize}
\item we use the objects from $\mathbf{X}$ as objects of $\mathcal{X}$: $\Ob(\mathcal{X}) = \mathbf{X}$;
\item the pair of objects $x_1$ and $x_2$ from $\Ob(\mathcal{X})$ is connected by a morphism $\alpha \in H_\mathcal{X}(x_1, x_2)$, if and only if there exists a transformation $a \in \mathbf{A}$, such that $x_2 = a(x_1)$;
\item the composition of morphisms in $\mathcal{X}$ is consistent with the composition of transformations from $\mathbf{A}$.
\end{itemize}

Our main object of interest is, however, the latent space $\mathbf{Z}$ of our model, since we suppose that each object from $\mathbf{X}$ is generated by some hidden variable $z \in \mathbf{Z}$.
Let's introduce a category $\mathcal{Z}$ in a following way:
\begin{itemize}
\item we use the objects from $\mathbf{Z}$ as objects of $\mathcal{Z}$;

\item we say, that there is a morphism $a': Z_1 \to Z_2$, if there is a morphism $a: X_1 \to X_2$. We denote this new set of morphisms $Hom(\mathcal{Z}) = \mathbf{A}'$.
\end{itemize}

Let's also denote the embedding, learned by our model, by $P : \mathbf{X} \to \mathbf{Z}$. 

\begin{figure}
    \centering
    
    \begin{tikzpicture}[node distance=2cm, auto]
    \node (X1) {$X$};
    \node (X2) [right of=X1] {$X$};
    \node (Z1) [below of=X1] {$Z$};
    \node (Z2) [below of=X2] {$Z$};
      
    \draw [decoration={markings,mark=at position 1 with
    {\arrow[scale=\arrowheadsizediag,>=stealth]{>}}},
    postaction={decorate}] (X1) -- node [above=0.5mm] {$a$} (X2);
    \draw [decoration={markings,mark=at position 1 with
    {\arrow[scale=\arrowheadsizediag,>=stealth]{>}}},
    postaction={decorate}] (Z1) -- node [above=0.5mm] {$a'$} (Z2);
    
    \draw [decoration={markings,mark=at position 1 with
    {\arrow[scale=\arrowheadsizediag,>=stealth]{>}}},
    postaction={decorate}] (X1) -- node [left=0.5mm] {$P$} (Z1);
    \draw [decoration={markings,mark=at position 1 with
    {\arrow[scale=\arrowheadsizediag,>=stealth]{>}}},
    postaction={decorate}] (X2) -- node [right=0.5mm] {$P$} (Z2);
    
    \draw [dashed, decoration={markings,mark=at position 1 with
    {\arrow[scale=\arrowheadsizediag,>=stealth]{>}}},
    postaction={decorate}] (X1) -- node [left=0.5mm] {} (Z2);
    \end{tikzpicture}
    
    \caption{Diagram we want to be commutative}
    \label{commutativity}
\end{figure}
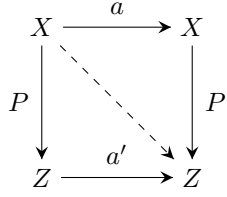

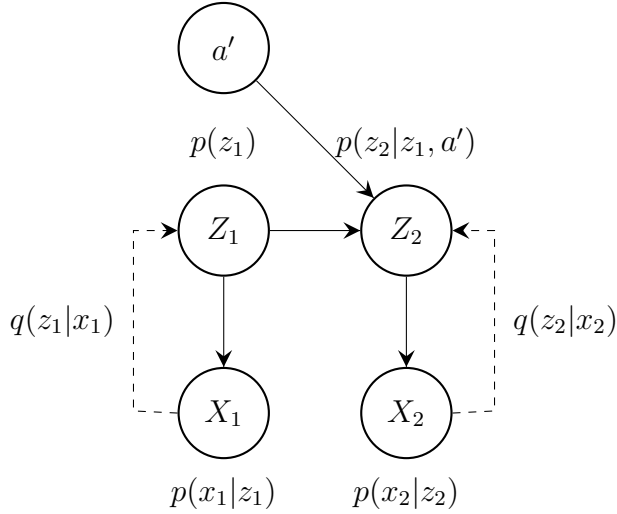
\begin{figure}
    \centering    
    \nodefontsize
      
    \begin{tikzpicture}[
    node distance=\nodedistance mm,
    nonterminal/.style={
    	circle,minimum size=\nodesize mm,
    	thick,draw=black,font=\ttfamily},
    terminal/.style={
    	circle,minimum size=\nodesize mm,
    	thick,draw=black,font=\ttfamily}]
    
    \node (Z1) [nonterminal] {$Z_1$};
    \node (X1) [nonterminal,below=of Z1] {$X_1$};
    \node (Z2) [nonterminal, right=of Z1] {$Z_2$};
    \node (X2) [nonterminal,below=of Z2] {$X_2$};
    
    \node (a)  [terminal,above=of Z1] {$a'$};
    
    \draw [decoration={markings,mark=at position 1 with
    {\arrow[scale=\arrowheadsize,>=stealth]{>}}},
    postaction={decorate}] node [above=0.8cm] {$p(z_1)$} (Z1) -- (X1) node [below=0.7cm] {$p(x_1|z_1)$} ;
    
    \draw [decoration={markings,mark=at position 1 with
    {\arrow[scale=\arrowheadsize,>=stealth]{>}}},
    postaction={decorate}] (Z2) -- (X2) node [below=0.7cm] {$p(x_2|z_2)$};
    
    \draw [decoration={markings,mark=at position 1 with
    {\arrow[scale=\arrowheadsize,>=stealth]{>}}},
    postaction={decorate}] (Z1) -- node [above=0.1cm] {} (Z2);
    
    \draw [decoration={markings,mark=at position 1 with
    {\arrow[scale=\arrowheadsize,>=stealth]{>}}},
    postaction={decorate}] (a) --  (Z2) node [above=0.8cm] {$p(z_2|z_1,a')$};
    
    \draw [dashed, decoration={markings,mark=at position 1 with
    {\arrow[scale=\arrowheadsize,>=stealth]{>}}},
    postaction={decorate}] (X1.west) -- (-\nodesize/10, -\nodesize/10-\nodedistance/10) -- node [left=0.1cm] {$q(z_1|x_1)$} (-\nodesize/10, 0) --  (Z1.west);
    
    \draw [dashed, decoration={markings,mark=at position 1 with
    {\arrow[scale=\arrowheadsize,>=stealth]{>}}},
    postaction={decorate}] (X2.east) -- (\nodedistance/10 + \nodesize * 2 / 10, -\nodesize/10 -\nodedistance/10) -- node [right=0.1cm] {$q(z_2|x_2)$} (\nodedistance/10 + \nodesize * 2 / 10, 0) -- (Z2.east);
    \end{tikzpicture}
    
    \caption{Graph model}
    \label{graph}
    \normalsize
\end{figure}

Now, let's consider the diagram on figure \ref{commutativity}. The goal of our model in the language of category theory is to make this diagram commutative. I.e.~for each $x_1 \in \mathbf{X}, a \in \mathbf{A}$, we want the existence of $a' \in \mathbf{A}'$, such that the equality $a'(P(x_1)) = P(a(x_1))$ holds. To achieve this we are using graph model, depicted on figure \ref{graph}.

\subsection{Variational Inference}

Let us first remind the main steps of the variational inference for the classical Variational Autoencoder \cite{VAE} and then derive it for our modified model.

Recall the identity (see, for instance, \cite{tutorial_on_vae})
\begin{equation}
\label{elbo_identity}
     \expect_{z\sim q(z|x)} [\log p(x|z)] - \KL{q(z|x)}{p(z)} = \mathcal{L}(x),
\end{equation}
which holds for any distribution $q(z|x)$ over the same domain $Z$ as $p(z|x)$. Here
\begin{equation}
\label{elbo_definition}
    \mathcal{L}(x) = \log p(x) - \KL{q(z|x)}{p(z|x)}
\end{equation}
is the evidence lower bound (ELBO) and
\begin{equation}
    \KL{p_1(z)}{p_2(z)}= \expect_{z\sim p_1(z)} \log\frac{p_1(z)}{p_2(z)}
\end{equation}
is the Kullback–Leibler divergence between two distributions. Since the latter is always non-negative one can obtain the following lower bound of the evidence from definition (\ref{elbo_definition}):
\begin{equation}
    \log p(x) \geq \mathcal{L}(x).
\end{equation}
Moreover, according to (\ref{elbo_identity}), in order to maximize the data likelihood (with respect to some trainable parameters $\theta$ hidden in $p(x|z)$ and $q(z|x)$) one has to maximize objective function $\mathcal{L}(x)$ along with minimizing the KL-divergence between $q(z|x)$ and $p(z|x)$. As usual, we assume that our encoding model $q(z|x)$ is complex enough to nullify the divergence term without affecting the log-likelihood term, so it suffices just to maximize $\mathcal{L}(x)$.

In our case the space $X$ of all observable variables consists of triples $(x_1,x_2,a) \in \mathbf{X}\times\mathbf{X}\times\mathbf{A}$ of two images and a transformation. We want our generative process to produce triples for which $x_2=a(x_1)$ more likely than other, thus we want to maximize $p(x_1,x_2,a)$ for these points. The space $Z$ of all latent variables consists of triples $(z_1,z_2,a') \in \mathbf{Z}\times\mathbf{Z}\times\mathbf{A}$. Here $a'$ is a formal duplicate of observable variable $a$ in the latent space involved in the generative process. Recall that the objective of auxiliary encoding function $q(z|x)$ is to give us distribution over $z$ values that are likely to produce $x$, but its form is completely a matter of our choice. We use the following one:
\begin{equation}
\label{q_form}
    q(z_1,z_2,a'|x_1,x_2,a) = q(z_1|x_1) q(z_2|x_2) p(a'|a).
\end{equation}
The conditional distribution $p(a'|a)$ can be written in its explicit form $p(a'|a) = p(a|a') = \delta(a - a')$.

The equality that follows from our graph model depicted on figure \ref{graph} will also be useful:
\begin{equation}
\label{p_form}
    p(x_1,x_2,a|z_1,z_2,a') = p(x_1|z_1) p(x_2|z_2) p(a|a').
\end{equation}
Thus the whole generating process in our model is the following:
\begin{multline}
    p(x_1,x_2,a) = \expect_{z_1,z_2,a'\sim p(z_1,z_2,a')} p(x_1,x_2,a|z_1,z_2,a')\\
    = \expect_{z_1\sim p(z_1)} \expect_{z_2\sim p(z_2|z_1,a)} p(x_1|z_1) p(x_2|z_2) p(a).
\end{multline}

In order to obtain the computable form of the evidence lower bound to train the model one has to substitute (\ref{q_form}) and (\ref{p_form}) into the identity (\ref{elbo_identity}) (written for specified above observable and hidden variables):
\begin{multline}
\label{th_objective}
    \mathcal{L}(x_1,x_2,a) = \expect_{z_1,z_2,a'}[\log p(x_1,x_2,a|z_1,z_2,a')] - \KL{q(z_1,z_2,a'|x_1,x_2,a)}{p(z_1,z_2,a')}\\
    = \expect_{z_1}[\log p(x_1|z_1)] +  \expect_{z_2}[\log p(x_2|z_2)] +\log p(a) -\\
    - \KL{q(z_1|x_1)}{p(z_1)} - \expect_{z_1}\KL{q(z_2|x_2)}{p(z_2|z_1,a)}.
\end{multline}
Here expectation $\expect_{z_1,z_2,a'}$ is taken over distribution $q(z_1,z_2,a'|x_1,x_2,a)$ and expectations $\expect_{z_i}$ over distributions $q(z_i|x_i)$ respectively.

In our experiments we did not use naturally augmented dataset $\bar{\mathcal{D}} = \{(x_1^i,x_2^i,a^i)|i\in I\}$, but did the augmentation by picking a transformation $a$ from finite predefined set $A = \{a^i|i\in I'\}$. Thus, given sampled $z_i\sim q(z_i|x_i)$ and $a\sim p(a)=\mathcal{U}_A(a)$ for every data point $x \in \mathcal{D}$, the objective (\ref{th_objective}) to maximize takes the form
\begin{multline}
\label{objective}
    \mathcal{L}(x,a(x),a|z_1,z_2) = \log p(x|z_1) + \log p(a(x)|z_2) + \log p(a) -\\
    - \KL{q(z_1|x)}{p(z_1)} - \KL{q(z_2|a(x))}{p(z_2|z_1,a)}.
\end{multline}
Here constant (with respect to trainable parameters) term $\log p(a)$ does not affect the maximization and can be omitted. However, one has to keep it if the distribution over possible transformations $p(a)$ is trainable itself.

\subsection{Details of Implementation}

More specifically, our model consists of two CNN encoders and two CNN decoders with the common parameter set and one full-connected network, which learns transformations (i.e.~morphisms $a'$ from $A'$) in their latent space $\mathbf{Z}$. All these networks are learning together via gradient descent with Adam optimizer, maximizing variational lower estimate of data likelihood (\ref{objective}).

In our particular implementation we used the following set  $A$ of allowed transformations:
\begin{itemize}
\item{up, down, left and right shifts by $3$px and $6$px;}
\item{scaling with coefficients $1.15$ and $1.32$;}
\item{rotations by $12^\circ$ and $24 ^\circ$.}
\end{itemize}

We assigned unique number for each transformation from this list. After it, at each training step we pick an object $x_1$ from our dataset with a particular transformation $a$ from $\mathbf{A}$ and obtain object $x_2$ by applying $a$ to $x_1$. Next, we compute the output of variational autoencoder on both objects --- $x_1$ and $x_2$.
Simultaneously we estimate a probability $p(z_2|z_1, a)$ of the object $z_2$ given $z_1$ and its desirable transformation $a$ with fully-connected network, mentioned above.

\section{Experiment Results}

In this section we compare the proposed model with several standard unsupervised algorithms that allow one to obtain a compressed representation of data in the latent space. We use the achievable accuracy of a linear classifier in the latent space as a measure of disentanglement of the obtained representation. We perform all the experiments on the MNIST dataset.

\subsection{2-dimensional Latent Space}

\begin{figure}
    \centering
    \subfloat[The MNIST dataset in the latent space]{{
        \includegraphics[width=.45\textwidth]{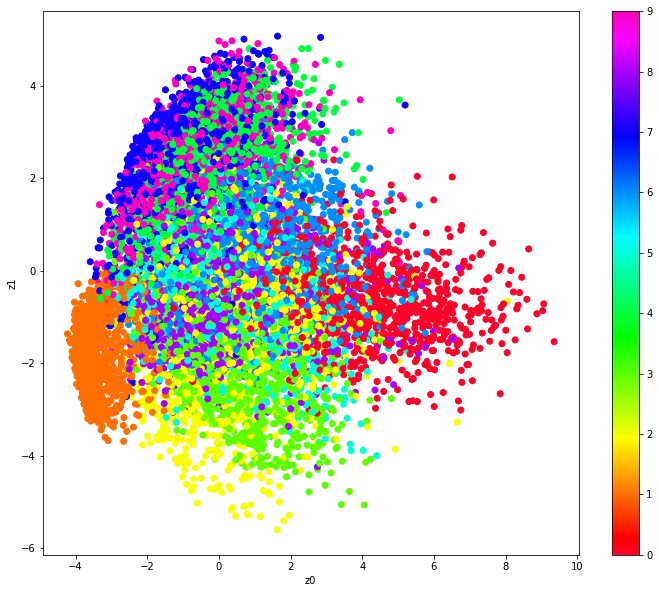}
    }}
    \qquad
    \subfloat[Orbit of each sample digit in the latent space]{{
        \includegraphics[width=.45\textwidth]{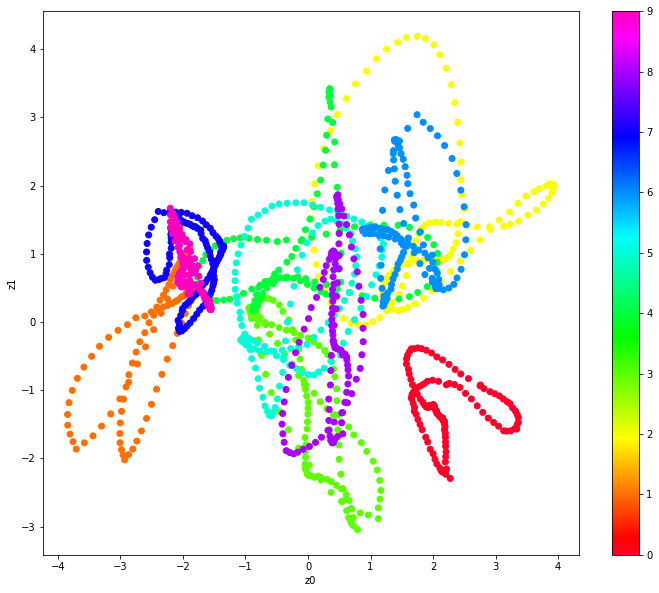}
    }}
    \caption{PCA embedding (first two components)}
    \label{pca_2dim}
\end{figure}

\begin{figure}
    \centering
    \subfloat[The MNIST dataset in the latent space]{{
        \includegraphics[width=.45\textwidth]{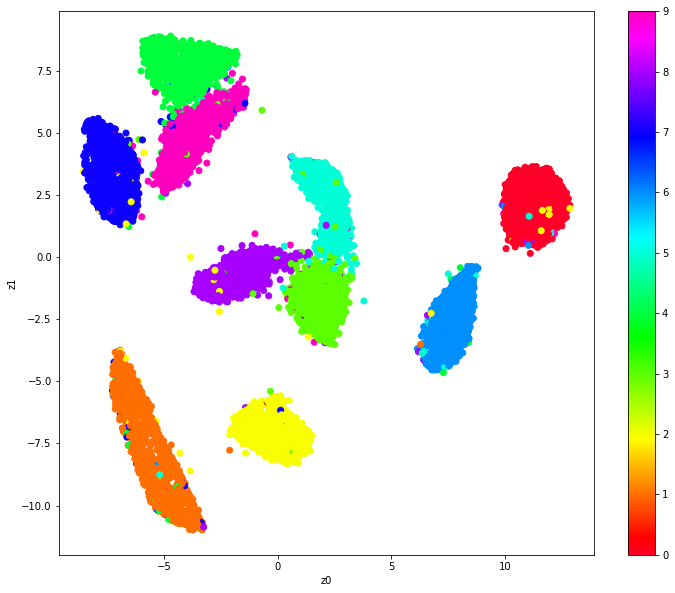}
    }}
    \qquad
    \subfloat[Orbit of each sample digit in the latent space]{{
        \includegraphics[width=.45\textwidth]{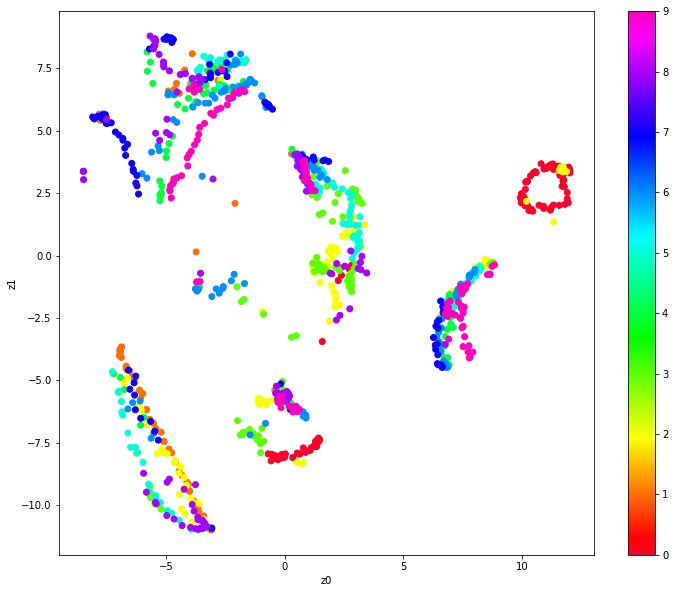}
    }}
    \caption{UMAP embedding (with default hyperparameters, see \cite{umap})}
    \label{umap_2dim}
\end{figure}

\begin{figure}
    \centering
    \subfloat[The MNIST dataset in the latent space]{{
        \includegraphics[width=.45\textwidth]{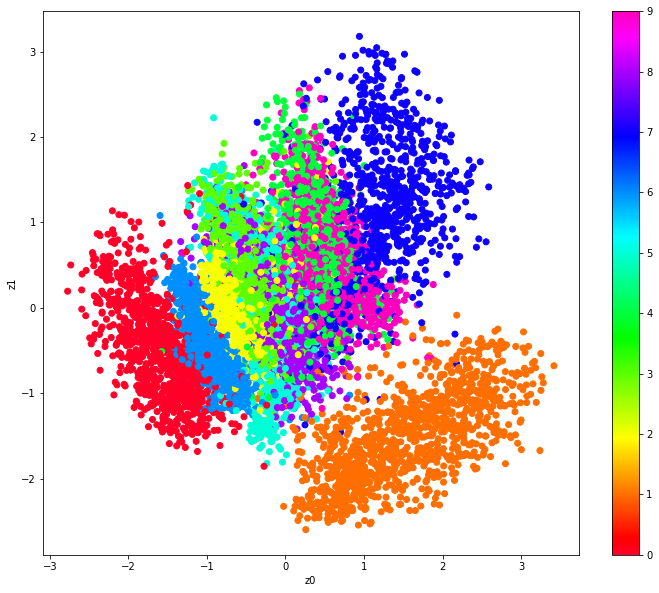}
    }}
    \qquad
    \subfloat[Orbit of each sample digit in the latent space]{{
        \includegraphics[width=.45\textwidth]{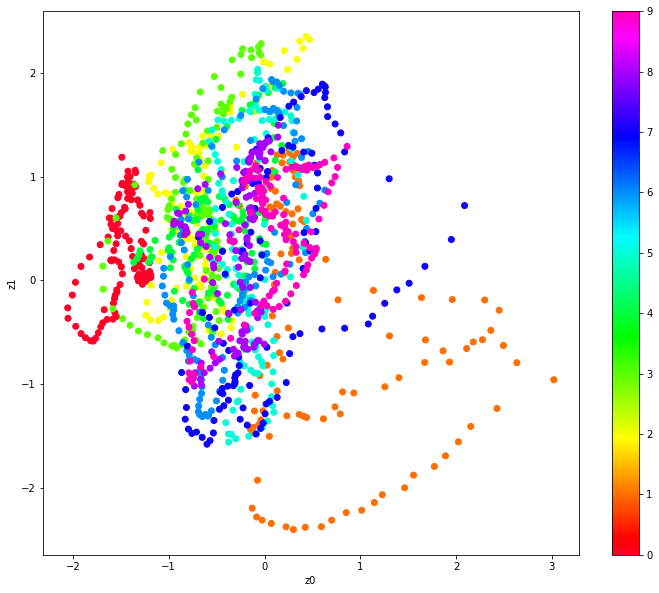}
    }}
    \caption{VAE embedding}
    \label{vae_2dim}
\end{figure}

\begin{figure}
    \centering
    \subfloat[The MNIST dataset in the latent space]{{
        \includegraphics[width=.45\textwidth]{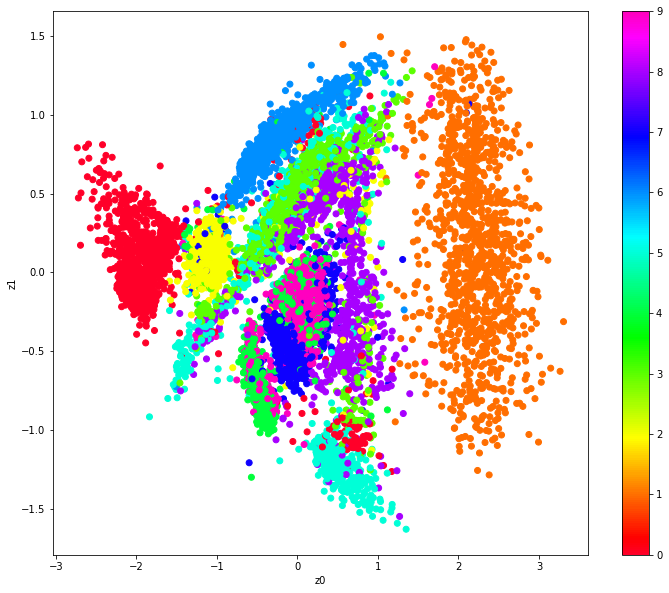}
    }}
    \qquad
    \subfloat[Orbit of each sample digit in the latent space]{{
        \includegraphics[width=.45\textwidth]{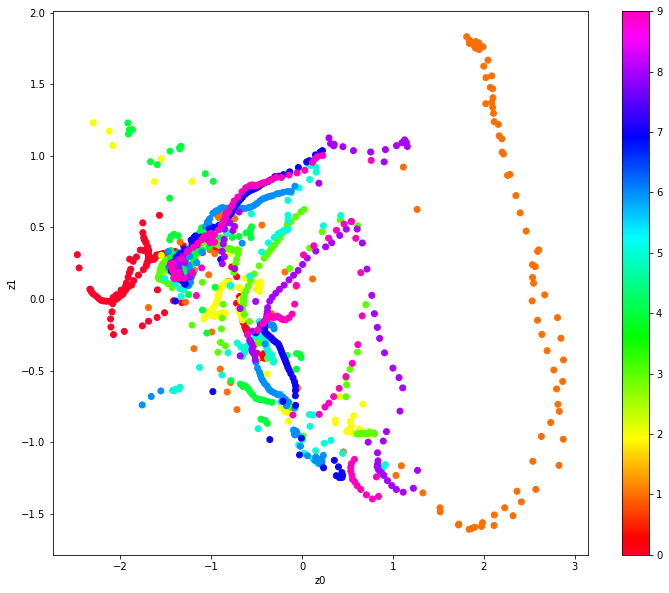}
    }}
    \caption{Our embedding (with hyperparameters $\beta = 0.01,\, \gamma = 5.0$)}
    \label{our_2dim}
\end{figure}

To demonstrate the distinctive features of the used algorithms we display the MNIST dataset embedded in 2-dimensional latent space by each of them (figures \ref{pca_2dim}--\ref{our_2dim}). To explore stability of such representation we also encode a sample of each of the ten handwritten digits from the dataset rotated $360^\circ$ with the step $3^\circ$. We call the obtained trajectories the orbits of these digits in the latent space.

\begin{figure}[H]
    \centering
    \begin{tabular}{||c|c||}  \hline
         Algorithm & Linear separability \\  \hline \hline
         PCA & 44.7\% \\ \hline
         UMAP & 95.5\% \\ \hline
         VAE & 63.0\% \\ \hline
         Our & 65.7\% \\  \hline
    \end{tabular}
    \caption{Accuracy of a linear classifier on the 2-dimensional latent space of different algorithms}
    \label{results_2dim}
\end{figure}

One can see measured accuracy on the testing set of a linear classifier trained on the training set embedded in the latent representation by each algorithm in the figure \ref{results_2dim}. Our algorithm along with classical VAE is not well-suited for data visualization task, thus it is outperformed by UMAP, which is.

\subsection{64-dimensional Latent Space}

\begin{figure}
    \centering
    \subfloat[The MNIST dataset]{{
        \includegraphics[width=.45\textwidth]{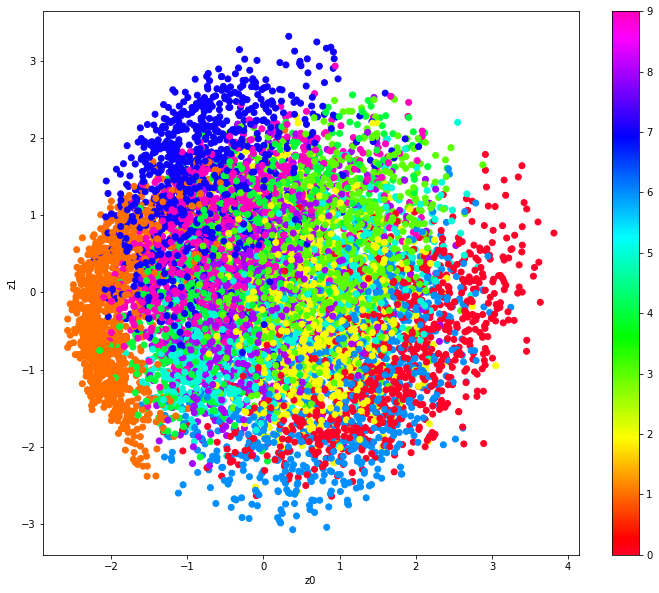}
    }}
    \qquad
    \subfloat[Orbit of each sample digit]{{
        \includegraphics[width=.45\textwidth]{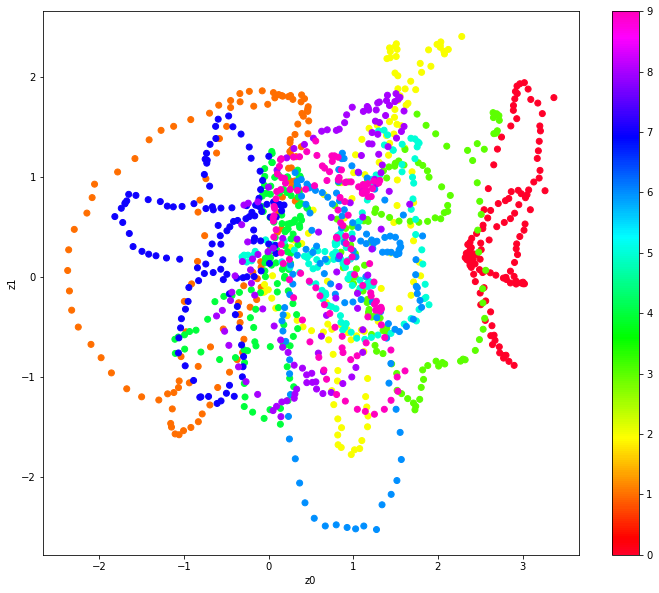}
    }}
    \caption{First two principal components projection of the 64-dimensional VAE embedding}
    \label{vae_64dim}
\end{figure}

\begin{figure}
    \centering
    \subfloat[The MNIST dataset]{{
        \includegraphics[width=.45\textwidth]{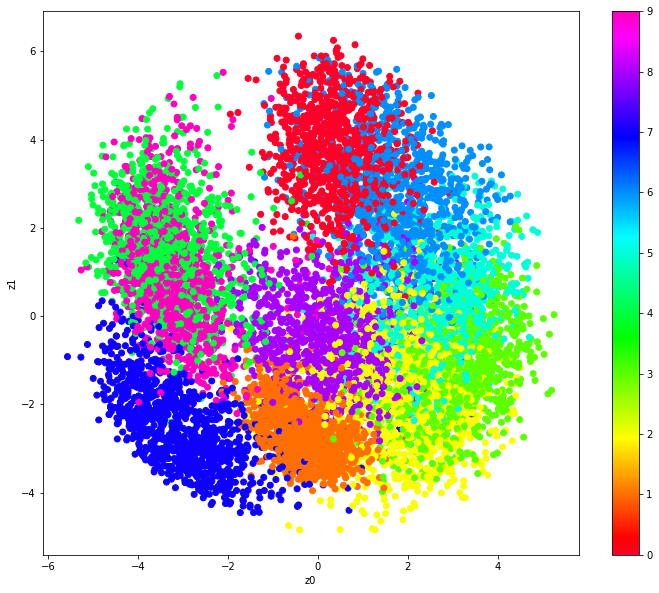}
    }}
    \qquad
    \subfloat[Orbit of each sample digit]{{
        \includegraphics[width=.45\textwidth]{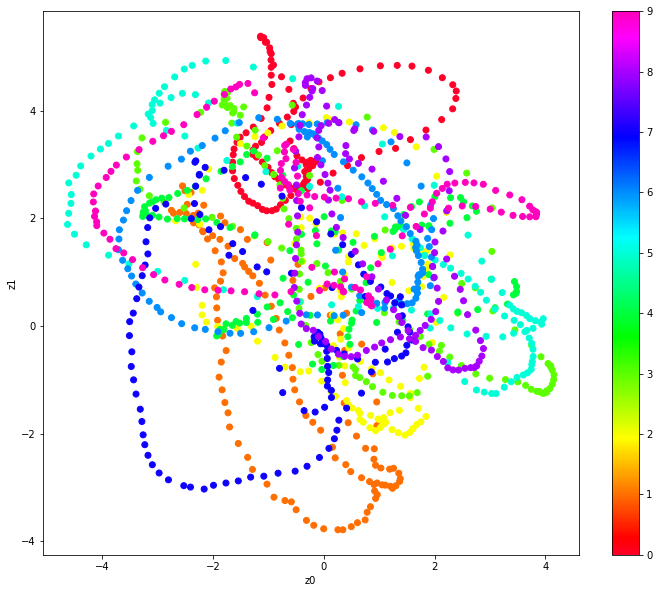}
    }}
    \caption{First two principal components projection of our 64-dimensional embedding}
    \label{our_64dim}
\end{figure}

\begin{figure}[H]
    \centering
    \begin{tabular}{||c|c||}  \hline
         Algorithm & Linear separability \\  \hline \hline
         PCA & 91.7\% \\ \hline
         UMAP & 96.0\% \\ \hline
         VAE & 96.2\% \\ \hline
         Our & 98.4\% \\  \hline
    \end{tabular}
    \caption{Accuracy of a linear classifier on the 64-dimensional latent space of different algorithms}
    \label{results_64dim}
\end{figure}

Once the latent space is large enough, our algorithm outperforms all the baselines and allows us to obtain better visualization (by projecting onto the first two principal components in the latent space). The smoothness and stability of the orbits indicates that the latent code changes well-predictable under geometric transformations.

\section{Conclusion and Future Work}

In this paper we have described a particular approach that allows a model to learn transformation-aware representation of real world objects without using labeled data. We have also shown that our model outperforms other well-known models in high-dimensional latent space in some aspects. We have provided mathematical generalization of the proposed principle, using the language of the Category theory, and described Bayesian inference for our model.

We believe that our ideas can be developed further in several directions and can be useful in other applications:

\begin{itemize}
    \item Being generalized to the wider class of transformations (beyond Gaussian), our model can learn to perform stochastic multi-modal transformations properly. Now if new objects appear on the scene after applying particular transformation, our model tends to predict many inconsistent intermediate states that look like partially appeared objects. One of the possibilities here is to adapt Generative Adversarial Network for our purposes, or use variations of VAE-like models with more flexible distribution families.
    
    \item Learning geometry of observed world and unsupervised scene understanding. Trained on a video sequence labeled with observer's movements, our model will require deep understanding of the scene in order to predict the next frame of the video. Such labeled observation sequences may be automatically generated and extracted from 3D simulations. In this class of tasks multi-modal probability models described above can be especially useful.
\end{itemize}

\end{document}